\documentclass{Interspeech}

\interspeechcameraready

\title{Automatic Speech Recognition Biases in Newcastle English: an Error Analysis}

\author[affiliation={1,2}]{Dana}{Serditova}
\author[affiliation={2}]{Kevin}{Tang}
\author[affiliation={3}]{Jochen}{Steffens}

\affiliation{English Department}{University of Freiburg}{Germany}
\affiliation{Faculty of Arts and Humanities}{Heinrich Heine University Düsseldorf}{Germany}
\affiliation{Institute of Sound and Vibration Engineering}{University of Applied Sciences Düsseldorf}{Germany}

\email{dana.serditova@anglistik.uni-freiburg.de, jochen.steffens@hs-duesseldorf.de, kevin.tang@hhu.de}

\usepackage{comment}
\usepackage{graphicx}
\usepackage{tipa}

\begin{document}

\maketitle

\begin{abstract}
    
   Automatic Speech Recognition (ASR) systems struggle with regional dialects due to biased training which favours mainstream varieties. While previous research has identified racial, age, and gender biases in ASR, regional bias remains underexamined. This study investigates ASR performance on Newcastle English, a well-documented regional dialect known to be challenging for ASR. A two-stage analysis was conducted: first, a manual error analysis on a subsample identified key phonological, lexical, and morphosyntactic errors behind ASR misrecognitions; second, a case study focused on the systematic analysis of ASR recognition of the regional pronouns ``yous'' and ``wor''. Results show that ASR errors directly correlate with regional dialectal features, while social factors play a lesser role in ASR mismatches. We advocate for greater dialectal diversity in ASR training data and highlight the value of sociolinguistic analysis in diagnosing and addressing regional biases.
\end{abstract}

\section{Introduction}

One of the major challenges of Automatic Speech Recognition (ASR) that has been repeatedly highlighted in research is social bias. While racial, age, and gender biases are well-documented \cite{koenecke2020racial, feng2024towards, liu2022towards, zellou2021age}, the investigations of regional biases remain limited \cite{wassink2022uneven, markl2022language, lai2024dialect}. This paper zooms in on ASR biases in Newcastle English, which is one of the most well-studied \cite{mearns2015tyneside, hughes2013english, schneider2004handbook}, yet one of the most demanding dialects for ASR systems \cite{markl2022language}.  

\subsection{ASR biases}

ASR systems struggle with language variation since they rely on acoustic and language models trained on limited data and mainstream dialects. Recent studies have inspected various adaptation strategies such as speaker-specific models, multi-dialect training, and accent identification, but there is still a gap in research with regard to dialectal and sociolectal variation \cite{wassink2022uneven, ngueajio2022hey, tatman2017effects}.

\subsubsection{Racial bias}

Research has demonstrated that existing ASR systems struggle with racial biases and exhibit higher Word Error Rate (WER) for marginalised communities. Koenecke et al. \cite{koenecke2020racial} found that Black speakers’ WERs were nearly double those of White speakers across ASR systems from major companies. Liu et al. \cite{liu2022towards} reported disparities in WER across skin tone groups, and Wassink et al. \cite{wassink2022uneven} found significant variation in WERs amongst Native American, African American, European American, and ChicanX speakers. Beyond phonetic differences, Martin \& Tang \cite{martin2020understanding} showed that morphosyntactic features that are absent in Mainstream Englishes, such as  ``habitual be'', a common AAE feature, increase WER in ASR.

\subsubsection{Age and gender bias}

Age and gender biases have been previously investigated and demonstrate a number of trends. Liu et al. \cite{liu2022towards} reported a consistent bias where male speakers receive higher WERs. The same gender-related result was demonstrated in Koenecke et al. \cite{koenecke2020racial}, with male speakers obtaining higher WERs compared to females. %
Similarly, Feng et al.\footnote{S. Feng, O. Kudina, B. M. Halpern, and O. Scharenborg, *Quantifying bias in automatic speech recognition*, arXiv:2103.15122, 2021.} found lower WERs for female speakers in a Dutch ASR system, with the largest difference observed amongst older adults and smallest amongst children. In terms of age, ASR performed best on teenage speech, and worst on children's speech, followed by older adults. %
However, the impact of age on ASR performance is not always consistent: age bias was not conclusive in Liu et al. \cite{liu2022towards}, with WERs remaining stable across the sample.

\subsubsection{Regional bias}

With more obvious social biases taking centre stage in current speech technology research, regional and dialectal biases often remain overlooked. By dialectal bias, we refer to systematic ASR errors that can be directly attributed to and explained by regionally specific linguistic features, including phonological, morphosyntactic, and lexical variation characteristic of a given dialect or speech community.

Research has highlighted ASR struggles with language variation and underrepresented dialects \cite{sanabria2023edinburgh}. Wassink et al. \cite{wassink2022uneven} found that sociophonetic variation contributes to ASR errors across ethnic dialects in the American Pacific Northwest. In the UK, ASR performance varies by region, with Southern British English speakers receiving the lowest WERs, while those from Northern England and Northern Ireland experience significantly higher error rates \cite{markl2022language}. The study suggests that regional biases stem from underrepresentation of these dialects in ASR training data and further reinforce existing linguistic marginalisation. Similarly, Torgbi et al.\footnote{M. Torgbi, A. Clayman, J. J. Speight, and H. T. Madabushi, *Adapting Whisper for Regional Dialects: Enhancing Public Services for Vulnerable Populations in the United Kingdom*, arXiv:2501.08502, 2025.} demonstrated that fine-tuning ASR on Scottish English improves recognition accuracy.

\subsubsection{ASR and sociolinguistic research}

Sociolinguistic insights can significantly improve ASR by revealing how dialectal variation drives recognition errors. Systematic error analysis can help identify problematic linguistic features that cause ASR errors. These insights can inform targeted data augmentation, fairness evaluations, and the development of more inclusive acoustic models that perform accurately across a greater range of speakers.

Wassink et al. \cite{wassink2022uneven} argue that current ASR systems struggle with dialectal variation because acoustic models are primarily trained on standard language varieties, while regional, ethnic, and social dialects are often unaccounted for. Markl \cite{markl2022language} makes a similar observation and states that quantitative evaluation alone is not sufficient to understand the causes of ASR errors. Instead, the need for qualitative and context-sensitive analysis that takes linguistic variation and social factors into consideration is emphasised. Torgbi et al.\textsuperscript{2} claim that WER is not a sufficient performance assessment metric and that ASR errors should be evaluated manually and qualitatively.

\subsection{Newcastle English}
\label{subsec:newcastle_english}

Newcastle English is one of the most recognised ones in England \cite{montgomery2012effect} and is well-researched by sociolinguists \cite{mearns2015tyneside, hughes2013english, schneider2004handbook}.  As demonstrated by Markl \cite{markl2022language}, it is also one of the most challenging UK accents for ASR. This study focuses on the most salient phonological, morphosyntactic, and lexical features of Newcastle English that might be particularly problematic for ASR.

Within phonetic and phonological features, the most salient ones include FOOT/STRUT split absence (e.g., ``fun'' [f\textipa{U}n]); lack of BATH retraction (e.g., ``path''  [pa\textipa{T}]); FACE and GOAT monophthongisation ([\textipa{e:}] and [\textipa{o:}]/[\textipa{8}]); PRICE vowel ([\textipa{EI}] or [\textipa{i:}]); NEAR variation (\textipa{[I@]} or \textipa{[E@]}); rounded NURSE vowel [\textipa{\o:}]; near glottalisation of /p, t, k/; `g'-dropping \cite{beal2004english, hughes2013english, mearns2015tyneside, docherty1999sociophonetic, grama2023post,watt2014patterns}. %

In terms of regional morphosyntactic features, the most prominent ones are unmarked plurals (e.g., ``six month" and ``three pound"); multiple negation; conflation of Past Tense and Past Participle forms (e.g., ``they’ve broke it"); regional pronouns ``yous'' (2nd pl.), ``wor'' (meaning ``our"), ``us'' in object position instead of ``me''; regional verb forms such as ``div'' and ``divn't'' (regional forms of ``do'' and ``don’t''), ``gan'' (``go'') \cite{beal1993grammar, beal2004english, pearce2012folk, hughes2013english, moelders2025navigating}.

\section{Method}

\subsection{Dataset}

To evaluate ASR performance on Newcastle English, we use the \textit{Diachronic Electronic Corpus of Tyneside English} (DECTE), which is a representative corpus of dialect speech from the Tyneside area of North-East England. It comprises 72 hours of naturalistic, spontaneous speech from a diverse range of 160 speakers (excluding the interviewers) from the Newcastle area, distributed across 99 files \cite{corrigan2012diachronic}. The corpus is fully transcribed by human annotators, which makes the output highly accurate and an ideal source for evaluating ASR performance.

\subsection{ASR system selection}

With a multitude of ASR systems on the market today, choosing a system to test requires careful consideration. We based our selection on two primary criteria: first, the system had to represent state-of-the-art ASR technology, incorporating the latest advancements in deep learning and speech processing; second, the system had to be commercially deployed and/or accessible to the average user, in order to ensure that our analysis reflects real-world application.

With these two factors in mind, we pre-tested the DECTE corpus on four ASR systems including both commercial solutions and more specialised open-source ASR: Google Cloud Speech-to-Text (\url{https://cloud.google.com/speech-to-text}), CrisperWhisper (an advanced variant of OpenAI's Whisper) \cite{zusag24_interspeech}, Deepgram Voice AI (\url{https://deepgram.com/}), and Rev AI (\url{https://rev.ai}).
10\% of the dataset were selected to evaluate these four ASR systems. Google Cloud Speech-to-Text and Deepgram Voice AI were eliminated first since they significantly underperformed compared to CrisperWhisper and Rev AI, with WERs ranging from 40\% to 60\%. CrisperWhisper and Rev AI were tested against each other on a bigger subset of files. Rev AI consistently outperformed CrisperWhisper across all speaker demographics, with a lower WER of approximately 10\% for most recordings. Across the entire dataset, Rev AI (set to English UK) achieved an average WER of 31.95\%.

\subsection{Data processing and analysis}

Data analysis consisted of two stages.\footnote{Data and code are available at: \url{https://doi.org/10.17605/OSF.IO/T8YXK}} First, a representative subsample of 16 DECTE files (32 speakers, with interviewers excluded from the analysis) transcribed by Rev AI was manually coded for errors. Errors were identified by comparing ASR output to human-checked transcriptions at word level. Two trained sociophoneticians (first author and an assistant) reviewed errors manually while listening to the audio. The analysis focused on dialectal bias and meaningful errors. By this we understand the types of errors that can be explained by dialectal variation and directly correlated with dialectal features in the region. The errors that were caused by noise, overlapping speech, or other indeterminate factors were excluded.

Errors were classified at two levels. First, by type: phonological, lexical, standardisation, morphosyntactic, and spelling. Standardisation errors involved replacing dialectal features with Standard Southern British English forms (e.g., ``me life'' → ``my life'', ``telly'' → ``television''). The remaining four error types correspond to distinct linguistic domains. The second level involved finer-grained classification within each error type. For instance, within phonological errors, we specified whether the errors were connected to vowel quality, monophthongisation, glottalisation, etc. For morphosyntactic errors, we defined them based on the grammatical category or syntactic phenomena they affected (pronouns, verb paradigm, tense and aspect, etc.).

Overall, 1,076 meaningful errors were included in the analysis. For each error, aside from the aforementioned classification, information on the gender and age of the speaker was included. This allowed us to conduct sociolinguistic analyses of errors and examine how social factors predetermine ASR performance.

In the second stage, we expanded our analysis to the full dataset (160 speakers) and examined errors that could be attributed to the local pronouns ``yous'' and ``wor'', since pronoun errors constituted one third of all morphosyntactic errors (see Sec.~\ref{subsubsec:morphosyntactic_errors}). We implemented an automated extraction process using custom Python scripts, which performed string alignment between the ASR output and human-verified transcriptions and identified mismatches where the target pronouns were misrecognised. The scripts then extracted each instance along with its surrounding context. Demographic information about the speaker was manually added for further sociolinguistic and statistical analysis.  
Given that noise has been shown to have impact on ASR inferences (e.g., \cite{martin2020understanding}), we estimated the Signal to Noise ratio (SNR) of each audio file using the Waveform Amplitude Distribution Analysis (WADA-SNR) \cite{kim2008robust}. To assess the effects of speaker age and gender on ASR performance, we fitted mixed-effects logistic regression models using \texttt{glmer} function from the \texttt{lmerTest} package \cite{Kuznetsova2017} in R (Version 4.4.1). Models were fitted separately for each pronoun with age and gender as fixed effects, speaker as a random effect, and noise level included as a covariate.

A key limitation is the absence of linguistic feature tagging, causing dialect patterns to be misclassified as errors. Incorporating feature tagging could provide a more accurate assessment of ASR performance on Newcastle English.

\section{Results}

The errors were analyzed with i) language-external variables, gender and age, and ii) language-internal variables consisting of error specifications and the relevant dialectal features.

\subsection{Language-external variables}
\label{subsec:language_external}

\begin{figure}[h!]
\begin{flushleft}
    \centering
    \includegraphics[width=0.48\textwidth]{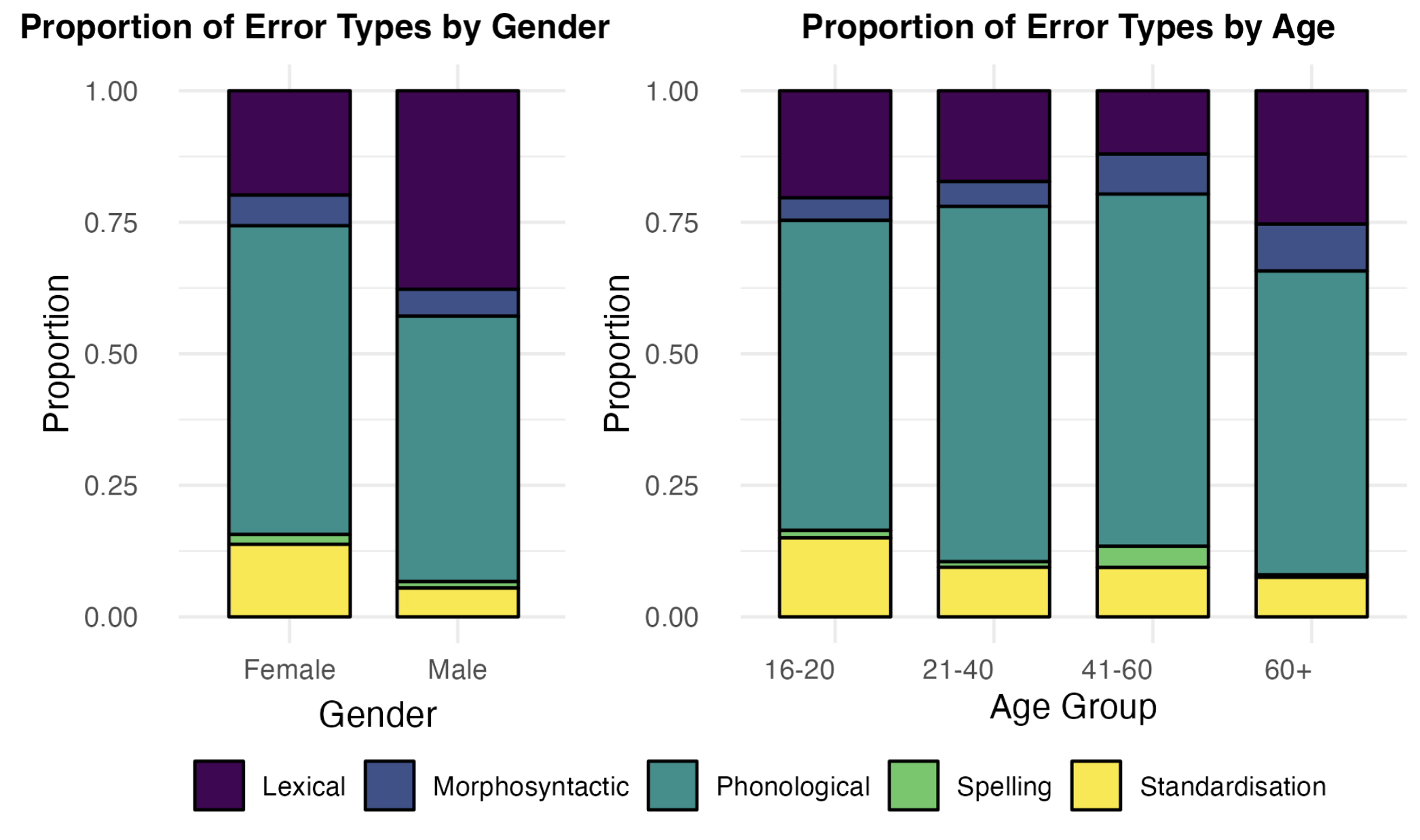}
    \caption{Proportion of error types by gender and age.}
    \label{fig:error_types_by_age}
\end{flushleft}
\end{figure}

Figure \ref{fig:error_types_by_age} (left)  shows the proportion of error types by gender. First, it is important to note that the male speakers in the dataset received higher error rates than their female counterparts (male speakers: 53.7\% vs. female speakers: 46.3\%, 2-sample test for equality of proportions: $\chi^2 = 11.6$, $p < 0.001$; overall distribution of errors: $\chi^2 = 5.948$, $p < 0.05$). As shown in Figure \ref{fig:error_types_by_age}, there are some gender differences in the error distribution. Male speakers received more lexical errors, e.g., multiple misrecognitions of the local lexical items ``nowt'', ``owt'', ``aye'', as well as the more common items like ``lass'', ``lads'', ``bloke'', ``quid'' (males: 37.02\% vs. females: 19.08\%, $\chi^2(1) = 41.224$, $p < 0.001$).

Morphosyntactic, phonological, and spelling errors were in relatively equal proportions for both genders (morphosyntactic: 5.62\% vs. 5.02\%, $\chi^2(1) = 0.093$, $p > 0.05$; phonological: 56.43\% vs. 49.48\%, $\chi^2(1) = 4.901$, $p < 0.05$; spelling: 1.81\% vs. 1.21\%, $\chi^2(1) = 0.306$, $p > 0.05$). However, female speakers in the sample were associated with higher rates of standardisation errors (13.25\% vs. 5.36\% for males, $\chi^2(1) = 19.351$, $p < 0.001$). These errors include ASR `correcting' the local pronouns (e.g., ``me dad'' → ``my dad'', ``meself'' → ``myself'', ``wor'' → ``our'' or ``my'', ``hisself'' → ``himself''). There were also many instances of replacing ``cause'' → ``because''. Some local syntactic features were standardised as well, e.g., unmarked plurals (``four month'' → ``four months'').

Figure \ref{fig:error_types_by_age} (right) shows the proportion of error types by age group. The most notable discrepancies appear in the lexical errors. Age groups 16-20 and 60+ received the highest number of lexical errors (19.3\% and 25\%, respectively), which visibly decrease for those between 41 and 60 (12\%). As mentioned above, many of these were associated with local lexical items such as ``nowt'', ``owt'', or ``wey''. Toponyms also constituted a large proportion of these errors. There were repeated misrecognitions of ``Sunderland'', ``North Shields'', ``Tyne'', and ``Toon''. Even the term ``Geordie'', which refers both to the local dialect and its speakers, consistently suffered from misrecognitions.\footnote{The lexical item ``aye'' is omitted in Figure \ref{fig:error_types_by_age}.  Since two men in the subsample used it persistently and frequently and it was misrecognised every time, it could overwhelm the sample and skew the result.}

\subsection{Language internal variables}

Here, we examine language-internal factors behind ASR errors, analyzing phonological, morphosyntactic, and lexical errors in relation to local dialectal features. Examples of standardisation errors were given in Sec.~\ref{subsec:language_external}. Spelling errors mostly consisted of spelling differences between British English and American English, where British spellings were substituted with an American alternative, and are not numerous.

\subsubsection{Lexical errors}

Unsurprisingly, lexical errors offer the most range and are particularly difficult to analyse quantitatively. We do not delve into details of these errors in the framework of this paper and offer only an overview. As expected, local lexical items presented the most difficulties for ASR. Amongst those, ``nowt'' and ``owt'' (meaning ``nothing'' and ``anything'') frequently posed challenges for transcription and were often mistranscribed as ``now'' or ``no'' and ``out''. Other local lexical features also led to errors, including ``lush'' (positive adjective), ``geet'' (emphatic ``very''), ``canny'' (versatile, typically positive), ``mint'' (``excellent''), and ``wey'' (another versatile expression that can mean ``well'', ``of course'', or used to express surprise).

\subsubsection{Phonological errors}

Phonological errors related to glottalisation, monophthongisation, and vowel quality were the most prominent. Near glottalisation of /t/ was the most challenging feature for ASR, causing nearly 23\% of phonological errors. These errors directly correlate with a well-documented feature of Newcastle English, where medial /p/, /t/, /k/ undergo glottalisation \cite{docherty1999sociophonetic, beal2004english, hughes2013english, mearns2015tyneside}. Examples of such errors are ``skipping'' [\textipa{skI\t{pP}In}] → ``skiing'' [\textipa{ski:IN}], ``patent'' [\textipa{peI\t{tP}@nt}] → ``paint'' [\textipa{peInt}], ``speaking up'' [\textipa{spi:\t{kP}In Up}] → ``speeding up'' [\textipa{spi:dIN Up}].

Monophthongisation of FACE [{e:}] accounted for 17.5\% of errors. This trend, led by young, middle-class females in Newcastle \cite{watt2014patterns, mearns2015tyneside} may challenge ASR systems trained on more conservative speech. This feature caused errors such as ``tray'' [\textipa{t\*r{e:}}] → ``tree'' [\textipa{t\*ri:}], ``whale'' [\textipa{w{e:}l}] → ``wheel'' [\textipa{wi:l}].
The third most common error was `g'-dropping [\textipa{In}] \cite{hughes2013english, beal2004english}, with over 11\% of errors caused by this feature. Examples of errors include ``drawing'' \textipa{[d\*ro:\textsci n]}  → ``drawn'' [\textipa{d\*rO:n}] and ``wedding'' [\textipa{wEdIn}] → ``wind'' [\textipa{wIn(d)}].

The remaining major error types are GOAT-monophthongisation (8.4\%), the glottalisation of /k/ and /p/ (6\% and 4.2\%, respectively), and glottal stops (4\%). Vowel quality was another major source of ASR errors (19.6\%), with FLEECE, NURSE, START vowels, as well as NEAR variation causing the most misrecognitions.

\subsubsection{Morphosyntactic errors}
\label{subsubsec:morphosyntactic_errors}

Local pronoun usage (38.6\%), verb paradigm (42.1\%), and tense and aspect (15.8\%) were most challenging morphosyntactic errors for ASR. As discussed in Sec.~\ref{subsec:newcastle_english}, Newcastle English is characterised by non-standard pronoun usage. Amongst these, the pronouns ``yous'' (2nd pers. pl.), ``wor'' (meaning ``our''), and 1st pers. sing. possessive ``me'' caused a third of all morphosyntactic errors (29.9\%). Verb paradigm, consisting of non-standard variation such as ``div'' and ``divnt'', first-person singular verb -s, and ``gan'', caused another third of errors in this category (35.1\%). Often, there was no output for these items (22\% of cases) or a phonetically similar item was used (e.g., ``didn't'', ``gun'', ``gone'') (13\% of cases). Finally, Past Perfect was persistently mistranscribed as Present Perfect by ASR, which can be explained by frequency bias (7\% of cases).

\subsection{ASR performance on local pronouns: yous and wor}

After conducting a qualitative analysis of all errors in the subsample and identifying recurring error patterns, we expanded our analysis to the full dataset. Here, we present a systematic analysis of errors affecting local pronouns ``yous'' and ``wor'', for which we implemented an automated extraction process to identify and quantify misrecognitions at scale.

These pronouns were selected for further analysis because they represent a salient morphosyntactic feature of Newcastle English that cannot be found in Standard Southern British English. Furthermore, these pronouns are frequently misrecognised according to our manual error analysis in Sec. \ref{subsubsec:morphosyntactic_errors}.

The results of the regression for the two pronouns are presented below. The models tested whether ASR produced the standardised forms (``yous'' → ``you'', ``wor'' → ``our'') or an incorrect output as the dependent variable.  For ``yous'', the model did not find a significant effect of age or gender ($\textit{p}s > 0.05$). However, the intercept was significant ($\hat{\beta} = -1.22$, $p < 0.01$), which indicates overall low recognition accuracy for this pronoun. 
For ``wor'', the model revealed a significant effect of age, with speakers aged 21–40 significantly more likely to be correctly recognised compared to the reference group 16-20 ($\hat{\beta} = 1.79$, $p < 0.05$). Additionally, higher noise levels were associated with significantly lower ASR accuracy ($\hat{\beta} = -0.13$, $p < 0.05$) (Table~\ref{tab:glmm_wor}).

\begin{table}[h!]
    \centering
    \caption{Regression summary for ASR Recognition of \textit{``wor''}}
    \begin{tabular}{lrcl}
        \hline
        \textbf{Fixed Effect} & \textbf{Estimate} & \textbf{Std. Error} & \textbf{p-value} \\
        \hline
        Intercept         & -0.3077  & 0.9156  & 0.7368      \\
        Age: 21--40       & 1.7853   & 0.8735  & 0.0410 *    \\
        Age: 41--60       & 1.4596   & 0.8310  & 0.0790     \\
        Age: 60+          & 2.4948   & 1.7916  & 0.1638      \\
        Gender: Male      & -1.6679  & 1.5082  & 0.2688      \\
        Noise (dB)  & -0.1279  & 0.0503  & 0.0111 * \\
        \hline
    \end{tabular}
    \vspace{0.5em}
    \small{Random effect: Speaker ($\sigma^2 = 0$).}
    \label{tab:glmm_wor}
\end{table}

\section{Discussion}

Overall, it is clear that the lexical and phonological features are the primary challenge for ASR and vary the most across genders and ages. The proportion of phonological errors is highest across all speaker demographics. Within these, errors that correlated with monophthongisation, glottalisation, and overall vowel quality were the most common. These errors should be studied systematically in future research. Methodologically, automating the annotation of dialectal features is a crucial step in examining these errors at scale \cite{Villarreal2020,Kendall_ing_2021}.

The higher proportion of lexical errors for men likely reflects their greater use of non-standard forms, which are associated with traditional masculinity and local identity. This is linked to covert prestige, where non-standard language forms carry social value within specific communities \cite{trudgill1972sex, labov1963social}. By contrast, research in sociolinguistics has consistently highlighted that women favour standard forms \cite{trudgill1972sex, labov1963social, milroy1980language, nichols1983linguistic}, and ASR models trained on mainstream data may reinforce these biases.

The variation in error proportions, particularly in the lexical errors, can be explained by age grading — a sociolinguistic process where speech patterns shift across life stages. Younger speakers frequently use more non-standard forms as linguistic innovators, while working adults tend to adopt standard forms due to professional and social pressures, and the influence of education. In later life, as these pressures decrease, speakers may return to vernacular features acquired in youth. Studies on Newcastle English confirm this pattern for `g'-dropping, the GOAT vowel, and possessive ``me'', and conclude that vernacular use peaks in adolescence, declines during adulthood, and increases again in retirement \cite{grama2023post, bauernfeind2023change, grama2023tracking, moelders2025navigating}.

The statistical analysis of ASR errors, particularly for the pronoun ``wor'', supports this claim. The results indicate that ASR struggles the least with this feature amongst working adults, while younger and older speakers receive more misrecognitions. These findings show that age grading interacts with ASR biases and that sociolinguistic patterns play a bigger role in ASR errors. Therefore, attributing these errors solely to model limitations is an oversimplification. 

It is also evident from our analyses that ASR errors are driven more by language-internal, dialect-specific factors than by language-external, social factors, despite some variation across speaker demographics. Thus, future research should address dialectal errors and acknowledge that social variables do not necessarily explain the full scope of ASR misrecognitions. 

\section{Conclusion}

The paper investigated ASR performance on Newcastle English and demonstrated that ASR errors can be directly attributed to local phonological, lexical, and morphosyntactic features. Through a manual error analysis that provided an overview of all error types, and a case study that looked into pronoun errors systematically, we established that language-internal, dialectal errors play a more substantial role in ASR mismatches than social factors. The results highlight the need for dialectal diversity in ASR training data \cite{Grieve2025} and the value of sociolinguistic insights in error assessment.

\section{Acknowledgements}

We would like to thank our research assistants, Paula Thees and Vincent Reichmann, for their support in data processing and analysis. We are also grateful to Karen Corrigan for granting us access to the DECTE corpus. This work was supported by the Hochschulinterne Forschungsförderung (HiFF) of the Hochschule Düsseldorf.

JS and KT are the senior authors. We follow the CRediT taxonomy\footnote{\url{https://credit.niso.org/}}. Conceptualisation: DS, KT, JS; Data curation: DS; Formal Analysis: DS; Funding acquisition: JS, KT, DS; Investigation: DS, KT; Methodology: DS, KT; Resources: KT, JS; Software: KT, DS; and Writing – original draft: DS, review \& editing: KT, JS.

\bibliographystyle{IEEEtran}
\bibliography{Tang}

\end{document}